\documentclass[sigconf]{acmart}
\usepackage{algorithm}
\usepackage{algorithmic}
\usepackage{pifont}
\usepackage{soul}
\usepackage{multirow}
\usepackage{makecell}

\copyrightyear{2024}
\acmYear{2024}
\setcopyright{rightsretained}
\acmConference[ICCAD '24]{IEEE/ACM International Conference on Computer-Aided Design}{October 27--31, 2024}{New York, NY, USA}
\acmBooktitle{IEEE/ACM International Conference on Computer-Aided Design
(ICCAD '24), October 27--31, 2024, New York, NY, USA}
\acmDOI{10.1145/3676536.3676826}
\acmISBN{979-8-4007-1077-3/24/10}
\begin{document}

\title{Spiking Transformer Hardware Accelerators in 3D Integration}

\author{Boxun Xu\textsuperscript{1}, Junyoung Hwang\textsuperscript{2}, Pruek Vanna-iampikul\textsuperscript{2, 3}, Sung Kyu Lim\textsuperscript{2}, Peng Li\textsuperscript{1}$^*$}
\email{{boxunxu, lip}@ucsb.edu, {jyh,v.pruek,limsk}@gatech.edu}
\affiliation{
  \textsuperscript{1}Department of Electrical and Computer Engineering, University of California, Santa Barbara, CA \country{USA} \\
  \textsuperscript{2}Department of Electrical and Computer Engineering, Georgia Institute of Technology, GA \country{USA} \\
    \textsuperscript{3}Department of Electrical Engineering, Burapha University, Chonburi \country{Thailand}
}


\begin{abstract}
Spiking neural networks (SNNs) are powerful models of spatiotemporal computation and are well suited for deployment on resource-constrained edge devices and neuromorphic hardware due to their low power consumption. Leveraging attention mechanisms similar to those found in their artificial neural network counterparts, recently emerged spiking transformers have showcased promising performance and efficiency by capitalizing on the binary nature of spiking operations. Recognizing the current lack of dedicated hardware support for spiking transformers, this paper presents the first work on 3D spiking transformer hardware architecture and design methodology. We present an architecture and physical design co-optimization approach tailored specifically for spiking transformers. 
Through memory-on-logic and logic-on-logic stacking enabled by 3D integration, we demonstrate significant energy and delay improvements compared to conventional 2D CMOS integration. 

\end{abstract}







\maketitle

\section{Introduction}

Transformer models have significantly advanced model capabilities in language modeling and computer vision, and have found widespread adoption across various application domains \cite{TransformerImageRec2021,transformerText2Image21}. 
At the heart of these models lies a self-attention mechanism,  which captures rich contextual information by considering all elements in a long input sequence, blending global and local sequence details into a unified representation.  

Spiking neural networks (SNNs) are more biologically plausible than their non-spiking artificial neural network (ANN) counterparts \citep{gerstner2002spiking}. Notably, SNNs can harness powerful temporal coding, facilitate spatiotemporal computation based on binary activations, and achieve ultra-low energy dissipation on dedicated neuromorphic hardware \citep{akopyan2015truenorth, davies2018loihi, PTB}.  Recent spiking transformers showcased promising performance and efficiency by capitalizing on the binary nature of spiking activation \cite{spikformer, spikformer_tracking, zhu2023spikegpt, yao2023spike}.


However, there is a current lack of dedicated hardware architectures for spiking transformers \cite{spikformer, spikformer_tracking, zhu2023spikegpt, yao2023spike}.  Our goal is to fill this gap by developing optimized architectures capable of accelerating spatiotemporal spiking workloads for spiking transformers using 3D integration as a technology enabler.  We see many opportunities that 3D integration can offer to enable biologically-inspired spiking transformers. Firstly, memory-on-logic stacking capability in 3D configurations allows for the storage of a significant portion of model parameters within local memory, ensuring swift and parallel memory access. Secondly, logic-on-logic stacking in 3D opens avenues for significant enhancements in energy efficiency, particularly in spike delivery management within SNN architecture.  Ultimately, in a longer run, the ultra-dense neuron-to-neuron connectivity enabled by 3D integration promises improvements in SNN learning accuracy and efficiency, thereby propelling semiconductor chip emulation closer to the capabilities of the human brain.

\noindent\textbf{Challenges and Contributions} In this work, we adopt face-to-face(F2F)-bonded 3D integration technology to enable dedicated spiking transformer accelerators with memory-on-logic and logic-on-logic configurations. \\
Contribution 1: We propose \textbf{the first} dedicated 3D accelerator architecture for spiking transformers, which explore spatial and temporal weight reuse to support spike-based computation in transformer models .\\
Contribution 2: We enable \textbf{the first} 3D memory-on-logic and logic-on-logic interconnection schemes to significantly minimize energy consumption and latency, whereby delivering  highly-efficient spiking neural computing systems with low area overhead. \\
Compared to 2D CMOS integration, the 3D accelerator offers substantial improvements. 
For the spiking MLP workload, it provides a 7.0\% increase in effective frequency, 50\% area reduction, and reductions of 7.8\% in power consumption, 68.3\% in memory access latency, and 69.5\% in memory access power. For the spiking self-attention workload, the enhancements include a 6.3\% increase in effective frequency, 50\% area reduction, and reductions of 1.5\% in power consumption, 74.2\% in memory access latency, and 49.3\% in memory access power.


\def\thefootnote{$*$}\footnotetext{Corresponding author.}

\section{Background}
\subsection{Spiking Neural Networks}
\textbf{LIF and IF Models.} The Leaky-Integrate-and-Fire(LIF) neuronal model is widely adopted in SNNs \cite{gerstner2002spiking}, which has the following discretized dynamics over time:
\begin{equation}\label{eqn_LIF_1}
V_{i}\left[t_{k}\right] = V_{i}\left[t_{k-1}\right] + \sum_{j\in RF}w_{ji}S_{j}[t_{k}] - V_{leak} \\
\end{equation}
\begin{equation}\label{eqn_LIF_2}
S_{i}\left[t_{k}\right] = \begin{cases} 
1 & \text{if  } V_{i}[t_{k}] > V_{th} \rightarrow V_{i}[t_{k}]=0\\
0 & \text{else          }                   \rightarrow V_{i}[t_{k}] = V_{i}[t_{k}]
\end{cases}
\end{equation}

The Integrate-and-Fire (IF) spiking neuron model, as a simplified spiking neuronal model, is commonly used in SNNs which are converted from a pretained ANN \cite{bu2023optimal, pmlr-v139-li21d}. The dynamics of the IF model is obtained by setting $V_{leak}$ in Equ.~\ref{eqn_LIF_1} to zero.

\begin{figure}[t]
    \centering
    \includegraphics[width=0.5\textwidth, clip, trim={2.7cm 3.3cm 3.5cm 2.6cm}]{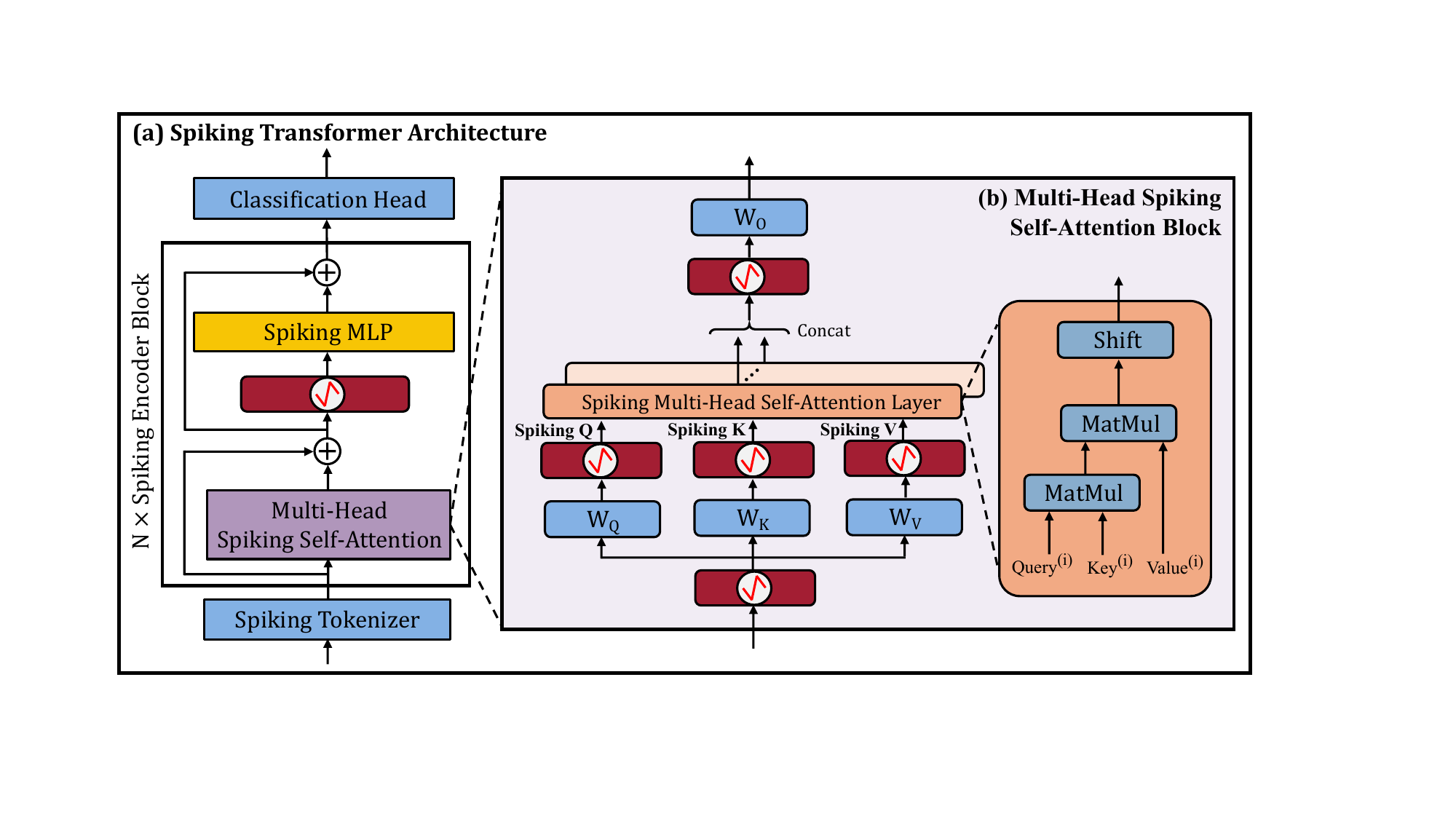}
    \caption{(a) Model Architecture of spiking transformers. (b) Multi-head spiking self-attention block within each spiking encoder block of spiking transformers.}
    \label{fig:spikingViT_arch}
\end{figure}

\noindent\textbf{Spiking Attention.}The spiking attention mechanism is proposed in \cite{spikformer} as the basic component of various variants of spiking-based transformers \cite{masked_spiking_transformer, spikformer_tracking}. The binary queries(Q) and keys(K) are correlated at each time point to compute the dependencies between tokens in attention maps; the binary values(V) are computed to reflect the attention-weighted accumulation for each token. Fig.~\ref{fig:spikingViT_arch} illustrates the architecture of spiking transformers.

\subsection{Neuromorphic Hardware}
\noindent\textbf{Neuromorphic inference/training accelerators.}
There exist various neuromorphic accelerators for SNN inference on the level of devices\cite{nebula}, circuits\cite{liu202430}, micro-architectures \cite{lee2018flexon},  architectures and on-chip communication networks \cite{davies2018loihi, debole2019truenorth, Spinalflow, PTB, Lee_ICCD_2020}; Some neuromorphic accelerators have been proposed for efficient SNN training \cite{skipper, yin2023sata, liang2021h2learn}. However, these accelerators are mainly tailored for spiking CNNs such as spiking AlexNet and ResNet \cite{fang2021deep, li2022converting, zheng2021going}. Although relevant to the acceleration of generic SNNs, these architectures are not optimized for large spiking transformers. 

\noindent\textbf{3D Neuromorphic hardware.}
Existing studies on the 3D IC realization of spiking neural networks have primarily employed monolithic 3D (M3D)\cite{M3D_SNN} and face-to-face (F2F) bonding techniques\cite{F2F_SNN}, which are rooted in traditional liquid state machines (LSM)-based architectures. While these works have adapted the M3D or F2F design methodologies to enhance power-performance-area (PPA) metrics, they exhibit compatibility issues with current spiking transformers and fail to provide optimized support for the latest advancements in spiking transformers, which have demonstrated competitive performance.
Furthermore, although the concept of memory-on-logic has been adopted in these work\cite{M3D_SNN, F2F_SNN}, they face limitations due to the restricted number of neurons, thereby constraining the potential for dataflow optimization. These limitations underscore the need for a more comprehensive approach that supports both logic-on-logic and memory-on-logic configurations, specifically tailored to enhance the functionality of spiking transformers.


\section{Proposed 3D Architecture Design}\label{sec: 3D_arch}
We introduce our proposed 3D dedicated architecture tailored for modern spiking transformers, which aims to enhance area utilization, energy efficiency, and processing speed.

In Section~\ref{sec: 3D_MLP}, we propose dedicated architecture support for managing the workloads of spiking MLP layers, which are key computational bottlenecks of transformers.  
Our proposed 3D dataflow minimizes data movement inherent in spiking transformers and maximizes weight reuse across tokens and timesteps, as well as the reuse of spiking activities across output features in MLP layers. 
Additionally, we design a dedicated systolic array that supports this optimized dataflow.
In Section~\ref{sec: 3D_ATTEN}, 
we introduce a dedicated 3D architecture tailored for the workloads of spiking attention layers, the other bottleneck of transformer models.  To address the challenges of heavy data movement associated with spiking attention maps, a kernel fusion strategy is employed to mitigate the need for extensive storage and data movement.
In Section~\ref{sec: 3D_REC}, 
we introduce a reconfigurable spiking self-attention array designed to flexibly handle various workloads of spiking attention layers, including key operations such as attention score computation $A=QK^T$ and $X=AV$. Our architecture and dataflow fully utilize fetched spiking query/key/value (Q/K/V)  and computed spiking attention score (A) data via maximized data reuse during execution. This approach not only minimizes area overhead but also reduces data movement, significantly enhancing both the efficiency and flexibility of the system.

\subsection{3D Acceleration for Spiking MLP layers} \label{sec: 3D_MLP}
\subsubsection{Workload Processing in MLP Layers}
MLP (linear) layers in spiking transformers encompass three core processing steps: \ding{192} synaptic integration, \ding{193} membrane potential accumulation, and \ding{194} spike generation, as detailed in Equ.~\ref{eqn_LIF_1} and Equ.~\ref{eqn_LIF_2}. 

\begin{figure*}[ht]
    \centering
    \includegraphics[width=\textwidth, clip, trim={3.2cm 1.7cm 7cm 2cm}]{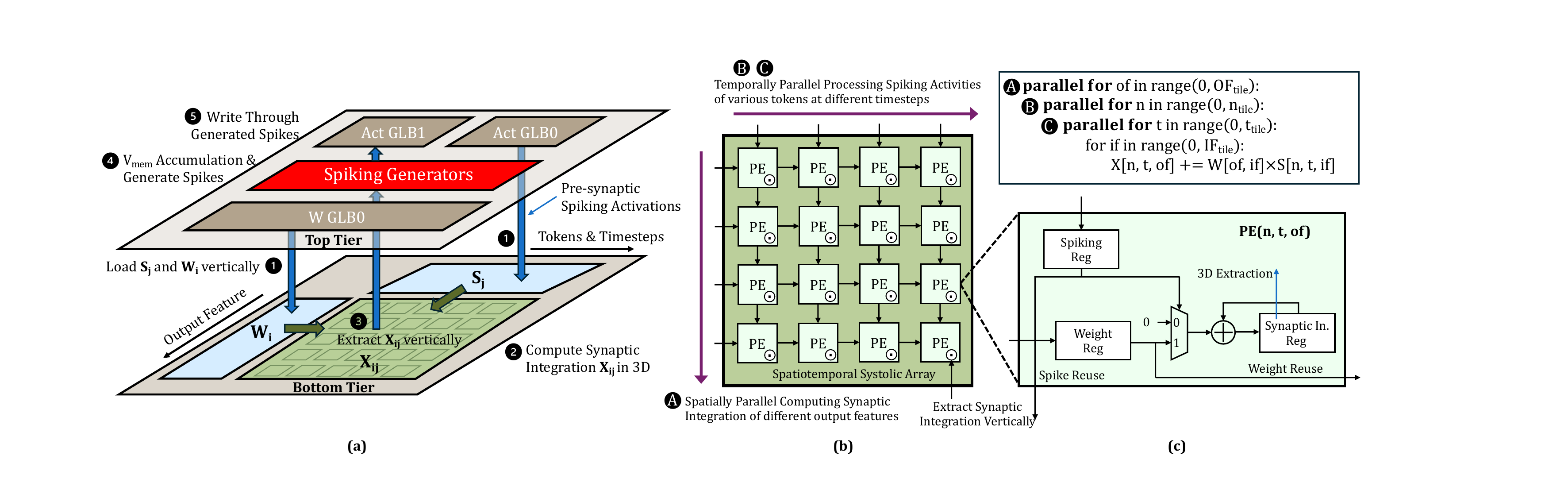}
    \caption{Proposed 3D Architecture for processing spiking MLP layers: (a) 3D partitioning and dataflow, (b) systolic PE array for synaptic integration on the bottom tier, and (c) PE design.  }
    \label{fig:3D_spiking_MLP}
\end{figure*}

The  step of \ding{192} is to process a pre-synaptic activation $S_{in}$ with a shape of $\mathbb{R}^{N\times T\times  D_{in}}$, using a pre-trained weight $W$ of shape $\mathbb{R}^{D_{in}\times D_{out}}$. Here, $N$ denotes the number of spiking tokens; $T$ represents the number of timesteps on which the model executes; $D_{in}$ and $D_{out}$ denote the number of input features and output features, respectively. As in typical SNNs \cite{fang2021deep}, $W$ can be shared for processing  all-or-none input spikes across multiple timesteps, based on which \cite{Lee_ICCD_2020, PTB} have proposed methods for executing such temporal workload in parallel. Unlike traditional spiking neural networks, spiking transformers are unique in the sense that weights can be further shared when processing different tokens from $S_{in}$ to gain additional benefits. While offering an outstanding opportunity, processing such complex workload both spatially and temporally is nontrivial, and requires a customized systematic design to exploit the potential data reuse  within the model.

The processing step of \ding{193} is to sequentially accumulate the computed synaptic integration of each neuron $i$ for each token $n$ at timestep $t$, denoted by $X_{n, i}[t]$, onto the membrane potential from the previous timestep, $V_{n, i}[t-1]$, to compute $V_{n, i}[t]$. Following this, step \ding{194} performs conditional spike generation at each timestep as outlined in Equ.~\ref{eqn_LIF_2}. This is done by comparing the current membrane potential $V_{n, i}[t]$ with the broadcasted voltage threshold $V_{th}$, determining whether an output spike shall be generated at each timestep.

Previous designs of neuromorphic accelerators typically provide limited parallelism, either in temporal or spatial dimensions, and are not tailored for spiking transformers. Additionally,  synaptic integration is hampered by a serial register readout chain, which introduces significant delays, or by complex wire routing, which complicates the extraction of computed integrated synaptic input at a given time point  and introduces high energy consumption. 

\subsubsection{Proposed 3D MLP Layer Architecture and Dataflow}
\begin{algorithm}[t]
\caption{Kernel Fusion for Spiking MLP Layers.}\label{alg:KF-S-MLP}
\begin{algorithmic}
\STATE {\bfseries Input:}  
    the number of token tiles $\#tile_{N}$,
    the number of time steps $\#tile_{T}$,
    the number of input features $\#tile_{IF}$,
    the number of output features $\#tile_{OF}$,
    the height/width of the systolic array $H/W$,
    spiking activation $S_{in} \in \mathbb{R}^{N\times T\times D_{in}}$,
    weight $W \in \mathbb{R}^{D_{in}\times D_{out}}$.
\STATE {\bfseries Output:}      
    spiking output $S_{out} \in R^{N\times T\times D_{out}}$.
\FOR{$of=0 \ \ \text{to}\ \#tile_{OF}-1$}
    \FOR{$n = 0 \ \ \text{to}\ \#tile_{N}-1$}
        \FOR{$t=0 \ \ \text{to}\ \#tile_{T}-1$}
            \FOR{$if=0 \ \ \text{to}\ \#tile_{IF}-1$}
                \IF {$W(of, if)$ not in W buf.}
                    \STATE \textbf{Load} $W(of,if)$ chunk from W GLB into W buf. 
                \ENDIF
                \STATE \textbf{Load} $S_{in}(n,t,if)$ chunk from Act GLB0 into Act buf.
                \STATE \textbf{Compute} synaptic integration chunk $X(n,t,of)$
            \ENDFOR
            \STATE \textbf{Extract} $X(n,t,of)$ to \textbf{Compute} $S_{out}(n,t,of)$ and \textbf{Write through} Act GLB1.
        \ENDFOR
    \ENDFOR
\ENDFOR
\end{algorithmic}
\end{algorithm}
In our proposed  3D integration with two silicon tiers as shown in Fig.~\ref{fig:3D_spiking_MLP}, we have a dedicated core systolic PE array core at the bottom tier to execute spiking kernel operation \ding{192}, and another dedicated Spiking Generator core at the top tier for the spiking kernel operation of \ding{193}+\ding{194}, respectively. 
As shown in the Alg.~\ref{alg:KF-S-MLP}, we adopt an efficient tiling scheme to enable kernel fusion of the above two spiking kernel operations. Each tile indexed by $of$, $n$, $t$ is loaded and processed sequentially; the computation between $W$ and $S_{in}$ tiles and spike generation computation of $S_{out}$ are operated in parallel.

Due to the significantly higher energy consumption and latency associated with data memory access compared to computation, most accelerators are designed to maximize the utilization of accessed data and enhance parallel computation\cite{PTB}. For spiking transformers given that a weight matrix is invariant with respect to various corresponding tokens and timesteps  in spiking MLP layers, it is feasible to implement weight reuse strategies for different tokens across different timesteps. Similarly, input spiking activities for a particular input feature can be reused across neurons that produce different output features. Enabling parallel execution across these three dimensions on hardware improves throughput without increasing data loading overhead.


As illustrated in Fig.~\ref{fig:3D_spiking_MLP}(a), the global memory buffers and spiking generators are placed on the top tier; the local buffers and systolic array are placed on the bottom tier. We design the following optimized dataflow. In step \ding{182}, a pre-synaptic spiking activation tile, $S_{in}$, and a weight tile, $W$ are vertically loaded from the global buffers Act GLB0 and W GLB at the top tier to the $S$  and $W$ buffer at the bottom tier. In step \ding{183}, synaptic integration of $OF_{tile}$ neurons for $n_{tile}$ tokens at $t_{tile}$ timesteps is computed within the spatiotemporal systolic array located at the bottom. In step \ding{184}, the computed synaptic integration is extracted vertically and fed into the spiking generators based on an appropriate time index. In step \ding{185}, the spiking generators compute the membrane potential and conditionally generate postsynaptic output spikes; in step \ding{186}, the spiking generators write through the generated spikes to the global buffer Act GLB1 at each timestep.

The bottom tier, shown in Fig.~\ref{fig:3D_spiking_MLP}(b), features a dense systolic array with  2D mapping of PEs. To leverage data reuse opportunities, spiking activities of different $n_{tile}$ tokens across different $t_{tile}$ timesteps are processed by PEs in different columns, and the spiking activities of different $OF_{tile}$ output features are handled by PEs in different rows. Between left-right-connected PEs, multi-bit weights across $OF_{tile}$ output features are propagated from left to right, being reused across different tokens and timesteps. Meanwhile, the input spiking activities $S_{in}$ are reused by different output neurons, by propagation from top to bottom.

Each PE, designed to be synaptic integration-stationary as shown in Fig.~\ref{fig:3D_spiking_MLP}(c), contains three registers serving as scratchpad memory. The registers store 1-bit input spiking activity, multi-bit weight, and multi-bit synaptic integration output, respectively. The accumulation of weight at the $if$-th input feature takes place only if the corresponding input spike is active. The synaptic integration output registers across the array are directly connected to the spiking generators at the top tier, leveraging the 3D extraction readout ports and high-density vertical wiring between the two silicon tiers. 

\subsection{3D Acceleration for Spiking Self-Attention Layers} \label{sec: 3D_ATTEN}
The computation of spiking self-attention layers is another bottleneck and encompasses several key operations: \ding{192} the computation of spiking attention maps ($A = QK^T$), \ding{193} attention-weighted synaptic integration ($X=AV$), which provides inputs to a set of LIF neurons for generating the final binary spike-based attention output, \ding{194} membrane potential accumulation of these LIF neurons, and \ding{195} conditional generation of the LIF neuron output spikes as the final attention output. In operation \ding{192}, the spiking query $Q$ and spiking key $K$, initially shaped as $\mathbb{R}^{T\times N\times D_{in}}$, are subdivided into $\mathbb{R}^{T\times N\times H\times d}$. Here, $T$ represents the number of timesteps; N denotes the number of tokens; $H$ and $d$ indicate the number of self-attention heads and the number of features per head, respectively. A spiking attention map $S \in \mathbb{R}^{T\times H\times N\times N}$ is computed for each head at each timestep. For instance, the spiking attention map at $t$-th timestep for $h$-th self-attention head results from the binary matrix multiplication of the spiking query and key at the specific head and timestep. 
In \ding{193}, the attention-weighted synaptic integration is executed for each head at each timestep. The spiking attention map $A$, serving as the attention weights, is combined with the spiking value $V$, shaped in $\mathbb{R}^{T\times N\times H\times d} $ to compute attention-weighted synaptic integration, denoted by $X$ shaped as $\mathbb{R}^{T\times N\times H\times d}$. 

\begin{figure*}[ht]
    \centering
    \includegraphics[width=\textwidth, clip, trim={0.5cm 0.4cm 7.2cm 2.4cm}]{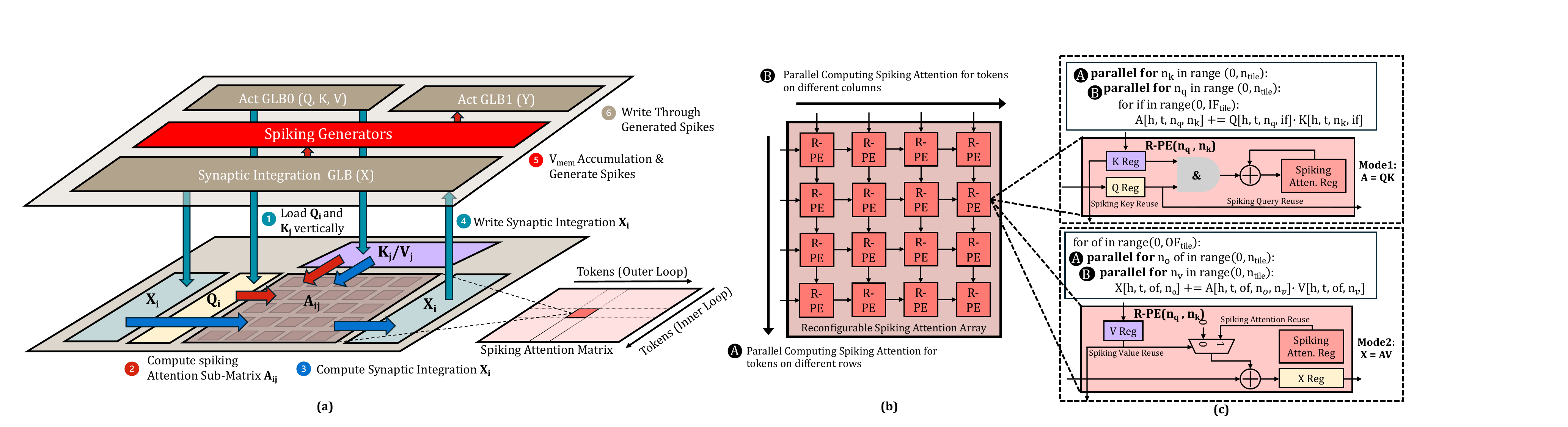}
    \caption{Proposed 3D Architecture for processing spiking attention layers: (a) 3D partitioning, (b) proposed reconfigurable dense systolic array on the bottom tier, and (c) reconfigurable PE design supporting two different computations.}
    \label{fig:3D_spiking_ATTEN}
\end{figure*}

Fig.~\ref{fig:3D_spiking_ATTEN} illustrates the dataflow for 3D integration based acceleration of spiking self-attention computation. In step \ding{182}, the partitioned spiking query $Q_{i}$ ($i$-th token) and key $K_{j}$ ($j$-th token) are vertically loaded from the global buffer activation GLB at the top tier to the Q buffers and K buffers at the bottom tier, respectively. In step \ding{183}, $Q_{i}$ and $K_{j}$ stream through the spiking self-attention array, detailed in Section~\ref{sec: 3D_REC}, where the multi-bit spiking attention map, $A_{ij}$, mapped to a submatrix of the whole spiking attention map $A$, is accumulated in an attention-stationary manner. Once this step is completed,  for computing $X = AV$, in step \ding{184} , the spiking value $V_{j}$ and partial synaptic integration $X_{i}$ vertically loaded into the V buffer and X buffer at the bottom tier are used to update synaptic integration. The attention-weighted $V$, that is the multiplication of $A_{ij}$ and $V_{j}$, is accumulated onto partial synaptic integration $X_{i}$. After the synaptic integration $X_{i}$ is computed, in step \ding{185}, it is written back to synaptic integration global buffer X GLB. In step \ding{186} and step \ding{187}, the spiking generators are activated to temporally accumulate the synaptic integration onto the membrane potential of each LIF neuron, and conditionally generate LIF neurons' spike outs, and write through the generated spikes to global buffer Act GLB 1, as the final spike-based outputs of spiking self-attention layers.


The space complexity of attention map is $O(H\times N^{2})$, which can be greater than the size of weight data with a space complexity of $O(D^{2})$, especially when dealing with multiple-framed videos and long linguistic contexts, where the number of tokens $N$ can be much greater than $D$. Furthermore,  the size of attention maps quadratically depends on $N$. Thus, reducing data movements of such huge attention maps is essential for efficient processing \cite{dao2022flashattention}.

We present a kernel fusion dataflow in spiking transformers, adapting from \cite{dao2022flashattention}, for this purpose 
as illustrated in Alg.~\ref{alg:KF-S-ATTN}.
In operation \ding{192}, a given spiking query token is reused while processing different spiking key tokens. Meanwhile, a spiking key token is also reused when processing different spiking query tokens. 
Similarly, in operation \ding{193}, each computed spiking attention element is reused when processing spiking values for different output features, and spiking values of a given output feature are reused while processing different spiking attention elements. Then, we design the dense array to enable the aforementioned two parallel processing schemes and two data reuse schemes for both operation \ding{192} and operation \ding{193}.

\begin{algorithm}[t]
\caption{Kernel Fusion for Spiking Attention}\label{alg:KF-S-ATTN}
\begin{algorithmic}
\STATE {\bfseries Input:}  
    the number of token tiles $\#tile_{N_k}$ and $\#tile_{N_q}$,
    the number of input features $\#tile_{IF}$,
    spiking query activation $Q \in \mathbb{R}^{T\times N\times H\times d}$,
    spiking key activation $K \in \mathbb{R}^{T\times N\times H\times d}$,
    spiking value activation $V \in \mathbb{R}^{T\times N\times H\times d}$.
    
\STATE {\bfseries Output:}      
    spiking output $S_{out} \in \mathbb{R}^{T\times N\times H\times d}$.
    \FOR{$h=0 \ \ \text{to}\ H-1$}
        \FOR{$t=0 \ \ \text{to}\ T-1$}
            \FOR{$i = 0 \ \ \text{to}\ \#tile_{N_k}-1$}
                \STATE \textbf{Load} $K(h,t,i)$ and $V(h,t,i)$ from ActGLB0 to K/V buf.
                \FOR{$j = 0 \ \ \text{to}\ \#tile_{N_q}-1$}
                    \STATE \textbf{Load} $Q(h,t,j)$ from Act GLB0 to Q buffer.
                    \STATE \textbf{Compute} $A(h,t,i,j)$ within the reconfigurable array.
                    \STATE \textbf{Load} partial synaptic integration $X(h,t,j)$ from X GLB to X buf.
                    \STATE \textbf{Compute} $X(h,t,j) = X(h,t,j) + A(t,i,j)\times V(t,i)$.
                    \STATE \textbf{Extract} partial synaptic integration $X(h,t,j)$ to X GLB. 
                \ENDFOR
            \ENDFOR
            \STATE \textbf{Compute} $S_{out}(h,t)$ and \textbf{write} Act GLB1. 
        \ENDFOR
    \ENDFOR
\end{algorithmic}
\end{algorithm}

\subsection{Reconfigurable Attention Array} \label{sec: 3D_REC}
To optimize dataflow and minimize data movement, we propose a reconfigurable spiking self-attention array that supports flexible matrix multiplication operations of $A=QK^T$ and $X=AV$. This design enables a flexible dataflow and reduces the frequency and volume of data movements associated with large, multi-bit attention matrices.

The proposed reconfigurable spiking self-attention array involves two modes. 

\textbf{Mode 1}: Each reconfigurable Processing Element (R-PE) computes one element of the spiking attention matrix. The spiking query is loaded from left to right, while the spiking key streams from top to down. The bit multiplication of spiking queries and keys is performed using a single two-input AND gate, with the results being accumulated onto a spiking attention register. Once computed, the attention map is stored within the array.

\textbf{Mode 2}: Each R-PE computes attention-weighted spiking value. If a propagated spiking value is active,  the stored attention in the R-PE will be accumulated onto synaptic integration. This partial synaptic integration is propagated from left to right, and the spiking value V is propagated from top to bottom, with the synaptic integration streaming out from the right boundary of the systolic R-PE array.

Each R-PE involves two 1-bit registers for storing Q and K/V, and two multi-bit registers to store spiking attention $A_{n_q, n_k}$ and synaptic integration, respectively.
Due to the binary nature of the spiking query, key, and value, we optimize the bitwidth to reduce redundancy by eliminating unnecessary high bit resolutions. The resolution required for all positive attention maps depends on the number of input features per head, necessitating a maximum of $log_{2}(d)+1$ bits. Additionally, the resolution for synaptic integration is determined by both the maximum number of input features and tokens, requiring up to $log_{2}(d)+log_{2}(N)+2$ bits. For an 8-head 128-feature spiking transformer with 128-token inputs, the bitwidth requirement of the attention map is $log_{2}(128/8)+1=5$ bits, and the bitwidth requirement of synaptic integration is 10 bits.

\section{Proposed 3D Physical Design}
\subsection{Memory on Logic}
To minimize the latency and energy consumption between memory and computing modules in spiking MLP layers and spiking attention layers of spiking transformers, we employ a memory-on-logic 3D stacking for 3D accelerators. In the spiking MLP accelerators, we group the spiking generators, spiking activation global buffers(Act GLB), and weight global buffer(W GLB) as a memory die on the top. The remaining components are organized as a logic die at the bottom. Similarly, in spiking self-attention accelerators, we group the spiking generators, activation global buffers(Act GLB), and synaptic integration global buffer (X GLB) as memory die, with other components placed on the logic die at the bottom. This configuration ensures balanced cell utilization between the top and bottom dies, thereby reducing latency and energy consumption of memory accesses to speed up the overall computation.

In the memory die of both accelerators, the activation global buffers, weight global buffer, and synaptic integration global buffer are implemented using SRAM to ensure its high density and compact footprint. Meanwhile, the remaining spiking generators are synthesized using the logic gates. 
In the logic die, the buffers adjacent to the systolic array also utilize SRAM to optimize space and efficiency, while other components are synthesized using logic gates during the logic synthesis phase. The placement of the SRAM modules is strategically pre-determined based on the data flow connections detailed in Section~\ref{sec: layout} for both MLP and attention layers. This strategic placement is designed to maximize the 3D connectivity between the memory and logic dies, enhancing both data transfer efficiency and overall system performance.

\subsection{Logic on Logic}

Unlike memory-on-logic stacking, logic-on-logic stacking provides enhanced flexibility in design space by allowing cell movement of standard cells on both dies. 
This flexibility supports various types of tier partitioning where the memory and logic areas are unbalanced. In the logic-on-logic stacking, we group the activation global buffers, weight global buffers, and spiking generators on the top, while the remaining compute logics are placed at the bottom. Similarly to the memory on logic stacking, we use SRAM for both activation global buffer and weight global buffer for high-density memory storage, and the remaining cells are synthesized with the combinational circuit to represent their functionality defined in the SystemVerilog. Therefore, the logic cells are placed alongside the memory macros and buffers are inserted in both top and bottom dies.
The logic-on-logic stacking enables the spiking generators to be connected with PEs with synaptic integration systolic array, and the spiking generators can extract the synaptic integration from bottom to top.

\begin{table*}[]
    \centering
\caption{The overall performance comparisons between 2D and 3D design of spiking MLP(linear) accelerator across different array size and bitwidth }
\begin{tabular}{c c c c c c c}
\toprule
\multirow{2}{*}{Array size H$\times$W, weight/synaptic integration bitwidth}   &   \multicolumn{2}{c}{$16\times 128$, 8b/16b}  & \multicolumn{2}{c}{$64\times 16$, 8b/16b} & \multicolumn{2}{c}{$64\times 16$, 4b/12b} \\  \cline{2-7}
                                        & 2D   & 3D         & 2D   & 3D     & 2D   & 3D    \\ 
\midrule
Effective Frequency ($GHz$) & 1.57 &  \textbf{1.68} &  1.68  &   \textbf{1.79}    &   1.76   & \textbf{1.85}               \\ 
Area Footprint ($mm^2$)  & 0.45$\times$0.9 & \textbf{0.45$\times$0.45} & 0.45$\times$0.78 & \textbf{0.45$\times$0.4} & 0.45$\times$0.78 & \textbf{0.45$\times$0.4} \\ 
Number of Cells & 152,335 & 152,012 & 88,838 & 88,447 & 83,931 & 83,923 \\ 
Wire length(m) & 1.37 & 1.10 & 1.17 & 0.99 & 1.00 & 0.81        \\ 
\midrule
Internal Power (mW) & 334 & 310 & 221.2 & 215.8 & 201.2 & 186.2 \\
Switching Power (mW) & 152 & 137 & 118.1 & 107.0 & 101.2 & 86.0 \\
Leakage Power (mW) & 30.0 & 29.1 & 22.8 & 21.0 & 19.1 & 14.4 \\
Total Power (mW) & 516 & \textbf{476.1} & 362.1 & \textbf{343.8} & 321.5 & \textbf{286.6} \\
\midrule
Memory Access Latency (ps) & 82 & \textbf{26} & 77 & \textbf{19} & 80 & \textbf{58}\\
Memory Access Power (mW) & 4.17 & \textbf{1.27} & 4.6 & \textbf{1.3} & 4.4 & \textbf{0.99} \\
\bottomrule
\end{tabular}    
    \label{tab:MLP_table}
\end{table*}

\subsection{Mapping 2D design to 3D}
With specific SRAM configuration to support the spiking generator buffers, we perform logic synthesis to obtain the initial gate-level netlist. The netlist is partitioned and grouped according to the specific hierarchy blocks for logic and memory groups. Following the tier partition, since the memory die contains the standard cells in both dies, we leverage the 3D design flow in \cite{PIN3D} to support the logic-on-logic physical implementation. This flow honors the partitioning information with two distinct cells for the top and bottom dies, where the pins are located according to the location of the die in the 3D metal stacks. The physical design (PD) stage is performed iteratively for one die at a time, while cells from another die are fixed. After the final step of the PD stage, we perform a static timing analysis (STA) to estimate the final Power, Performance, and Area (PPA). 

\begin{figure}[t]
    \centering
    \includegraphics[width=3.4in, clip, trim={1.7cm 1cm 2.2cm 0.6cm}]{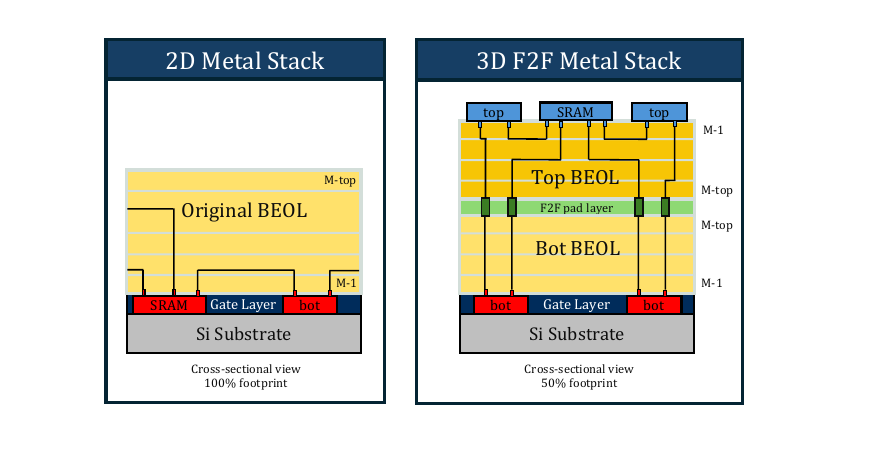}
    \caption{Cross-section view comparison between 2D and F2F 3D IC metal stack.}
    \label{fig:f2f_stack}
\end{figure}

\subsection{Face-to-Face Bonding}
Fig.~\ref{fig:f2f_stack} illustrates the difference between 2D and F2F 3D IC metal stack. To support the two-tier F2F 3D IC with \cite{PIN3D} design flow, we model the 3D interconnect and parasitics by combining two original 2D interconnect and connect them with vias to represent F2F bonding. We specify the via spacing to ensure that it meets the F2F pitch requirement. The F2F pitch is selected according to the grid size in the bonding layer.

In the case of the standard cells, we created two set of cells from original 2D cells for top and bottom die where pin layer are mapped into the 3D stacking. For memory macros, we changes their type to cover cells to allow standard cell placement of another die in the same region.

\begin{table}[]
    \centering
    \small
\caption{The overall performance comparisons between 2D and 3D design of spiking attention accelerator across different array size}
\setlength{\tabcolsep}{2.9pt}
\begin{tabular}{c c c c c}
\toprule
\multirow{2}{*}{Array size H$\times$W}   & \multicolumn{2}{c}{$16\times 16$}  & \multicolumn{2}{c}{$16 \times 8$}  \\ \cline{2-5}
                              & 2D   & 3D                          & 2D   & 3D    \\  \midrule
Effective Frequency ($GHz$) & 1.58 &  \textbf{1.68} &  1.68  &   \textbf{1.93}      \\ 
Area Footprint ($mm^2$)  & 0.45$\times$0.9 & \textbf{0.45$\times$0.45} & 0.45$\times$0.8 & \textbf{0.45$\times$0.4} \\ 
Number of Cells & 59,299 & 58,271 & 31,257 & 30,671 \\ 
Wire length(m) & 0.79 & 0.60 & 0.61 & 0.39      \\ 
\midrule
Internal Power (mW) & 146 & 146 & 97.8 & 95.8  \\
Switching Power (mW) & 52 & 51 & 30.6 & 26.3  \\
Leakage Power (mW) & 4.0 & 3.0 & 2.6 & 1.9  \\
Total Power (mW) & 203 & \textbf{200} & 130.9 & \textbf{124.0} \\
\midrule
Memory Access Latency (ps) & 388 & \textbf{100} & 388 & \textbf{72} \\
Memory Access Power (mW) & 3.22 & \textbf{1.63} & 3.86 & \textbf{1.36} \\
\bottomrule
\end{tabular}    
    \label{tab:ATTEN_table}
\end{table}

\section{Evaluations}
\subsection{Experiment Settings}
\textbf{Models, Datasets and Training Settings}
We evaluated the spiking transformer models trained on two widely adopted neuromorphic datasets: CIFAR10-DVS\cite{CIFAR10-DVS} and DVS-Gesture\cite{DVS128_gesture_dataset}. We adapted the same model setting using 4-bit quantized and 8-bit quantized results, respectively. Given that image sizes are uniform within a specific vision dataset, the token length of different samples remain consistent as in \cite{dosovitskiy2020image}. DVS-Gesture contains 11 hand gesture categories from 29 individuals under 3 illumination conditions; CIFAR10-DVS is a neuromorphic dataset containing dynamic spike streams captured by a dynamic vision sensor camera viewing moving images from the CIFAR10 datasets. In Tab.~\ref{table:dvs}, we demonstrate the superiority of the quantized spiking transformers over other SNNs, that can be efficiently executed on our proposed tiny 3D accelerators. The bitwidth in Tab.~\ref{table:dvs} indicates the bitwidth of spiking activation, synaptic weight and synaptic integration, respectively.

\begin{table}[h]
\centering

\begin{tabular}{ccccc}
\hline
\multirow{2}{*}{Model}  & \multicolumn{2}{c}{CIFAR10-DVS} & \multicolumn{2}{c}{DVS-Gesture} \\
\cline{2-5}
                                        & Bitwidth   & Acc.           & Bitwidth & Acc.        \\ \hline
Spiking VGG\citep{Fang_2021_ICCV}       & 1/32/32b       & 74.8\%       &1/32/32b & 97.6\%  \\
Spiking ResNet\citep{tdBN}              & 1/32/32b       & 67.8\%       &1/32/32b & 96.9\%   \\ \hline
\multirow{2}{*}{Spiking Transformer}    & 1/8/16b        & 81.2\%       &1/8/16b  & 98.26\%   \\
                                        & 1/4/12b        & 80.5\%       &1/4/12b  & 97.92\%   \\
\hline
\end{tabular}
    \caption{Comparison of the spiking transformer with other existing SNNs on CIFAR10-DVS and DVS-Gesture.}
\label{table:dvs}
\end{table}
\noindent\textbf{Hardware Platform Setup}
In this work, we use the commercial 28nm PDK to implement both 2D and 3D F2F designs. The 2D design consists of 6 metal layers, while the 3D design has double metal stack of 2D design with the F2F bond pitch varies from 0.5um to 1um. We use the Synopsys Design Compiler to synthesize the RTL to gate-level netlist and Cadence Innovus to perform physical synthesis.

For the memory, we utilize SRAM modules generated by a commercial memory compiler for various global buffers and storage functions within our system architecture. Specifically, 3072$\times$128b SRAM units are employed for the Activation Global Buffer (Act GLB), Weight Global Buffer (W GLB), and Synaptic Integration Global Buffer (X GLB), all placed on the top tier of our design.  Additionally, smaller 96$\times$128b SRAM macros are allocated for the Query (Q) buffer, Key/Value (K/V) buffer, and Spiking (S) buffer on the bottom tier. Two 96$\times$256b SRAM macros are configured to serve as extended X buffers.

\subsection{Overall Performance Comparision between 2D and 3D}

\subsubsection{Layout Comparision between 2D and 3D}\label{sec: layout} 
\begin{figure}[ht]
    \centering
    \includegraphics[width=0.5\textwidth, clip, trim={6cm 6cm 1.5cm 5.5cm}]{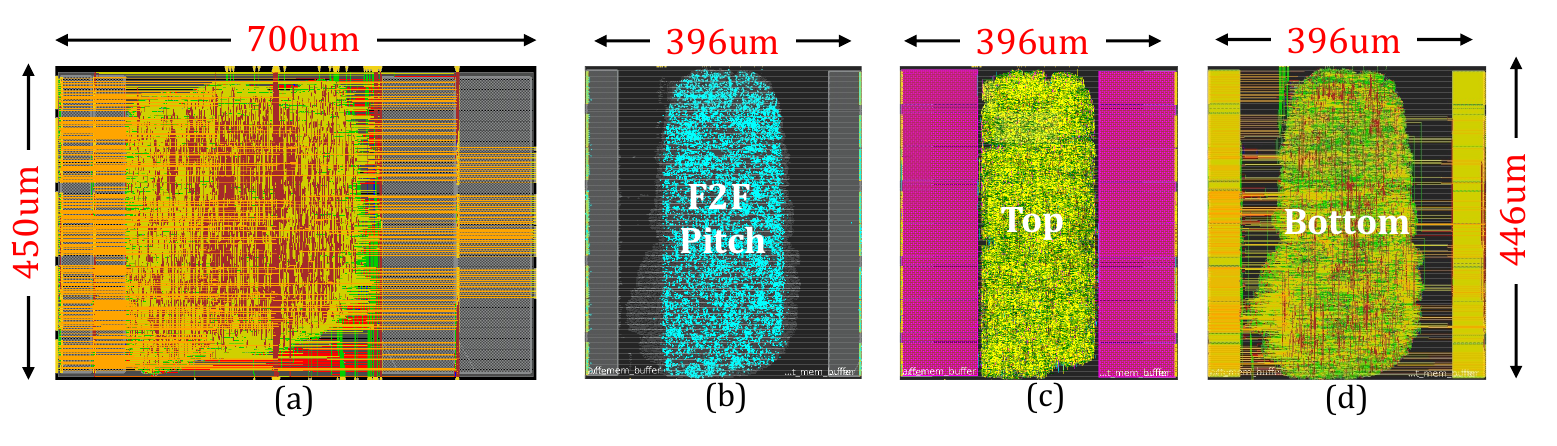}
    \caption{The layout comparison between 2D and 3D spiking MLP accelerators.(a)2D design. (b)(c)(d) 3D design.}
    \label{fig:MLP_layout}
\end{figure}

\begin{figure}[ht]
    \centering
    \includegraphics[width=0.5\textwidth, clip, trim={2.5cm 8cm 0.5cm 4.2cm}]{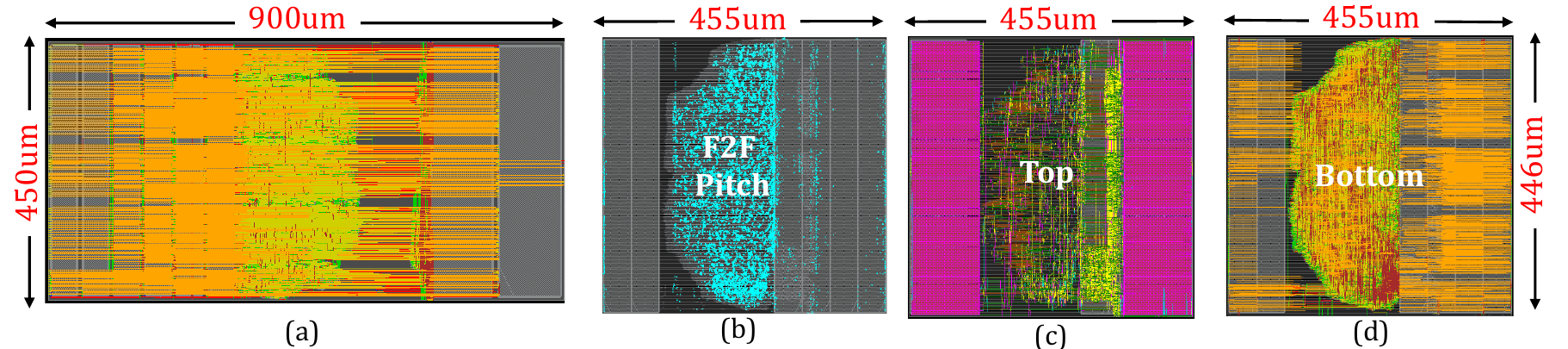}
    \caption{The layout comparison between 2D and 3D spiking self-attention accelerators.(a) 2D design. (b)(c)(d) 3D design.}
    \label{fig:atten_layout}
\end{figure}

In Fig.~\ref{fig:MLP_layout} and In Fig.~\ref{fig:MLP_layout} and Fig.~\ref{fig:atten_layout}, the layout is presented to show the difference between 2D and 3D design of spiking MLP accelerators and spiking self-attention accelerators. 

In Fig.~\ref{fig:MLP_layout}(a), the 2D design occupies $700um \times 450um$ while the stacked 3D spiking design occupies a $396um \times 446um$. On the top tier, the W GLB and Act GLB are placed on the edge in Fig.~\ref{fig:MLP_layout}(c), and the spiking generator array are occupied in the middle; on the bottom tier in Fig.~\ref{fig:MLP_layout}(d), the W and S buffers are placed on the edge, and the spiking spatiotemporal array is placed below the spiking generators. Fig.~\ref{fig:MLP_layout}(b), the F2F map indicates the interconnection between the top tier and the bottom tier.

In Fig.~\ref{fig:atten_layout}(a), the 2D design of spiking attention accelerator occupies $900um \times 450um$ while the stacked 3D spiking design occupies an area of $445um \times 446um$. On the top tier, the synaptic integration X GLB and Act GLB are placed on the edge in Fig.~\ref{fig:atten_layout}(c), and the spiking generator array is occupied in the middle; on the bottom tier in Fig.~\ref{fig:atten_layout}(d), the Q, K/V and X buffers are placed on the edge, and the spiking spatiotemporal array is placed between the buffers at the bottom tier. Fig.~\ref{fig:atten_layout}(b), the F2F map indicates the interconnection between the top tier and the bottom tier.

\subsubsection{PPA Comparision between 2D and 3D}

Tab.~\ref{tab:MLP_table} and Tab.~\ref{tab:ATTEN_table} demonstrate that 3D designs significantly enhance power, performance, and area efficiency compared to 2D designs. Using two dies in 3D designs allows the same cells to be accommodated within almost 50\% of the original area. This reduced physical distance between the spiking generator and accelerator array leads to shorter total wire lengths, reducing net delay and improving performance. The experiments showed an average 8\% increase in effective frequency. Furthermore, fewer buffers were used, decreasing the cell count and leading to a 6\% reduction in power consumption.

\subsubsection{3D improvement under different Array size}
Both the Spiking MLP accelerator and the Spiking Attention accelerator exhibit a decrease in effective frequency as the array size increases due to the increased number of cells, leading to higher routing congestion. To connect all these cells, ample routing resources are required, or the distance between cells must be minimized. In 2D designs, the spacing between cells is wider, and only a single layer of BEOL is utilized as a routing resource, necessitating longer wires. Conversely, 3D designs, which maintain narrower cell spacing and possess multiple metal layers as routing resources, demonstrate superior routing quality. As seen in Tab.~\ref{tab:aver_wire_length}, the wire length per net is significantly shorter in 3D designs. Longer wire lengths within a net can induce critical paths; thus, minimizing the distance between connected cells or increasing routing resources can enhance performance. Despite the decrease in effective frequency with increasing array size, as shown in Tab.~\ref{tab:MLP_table} and Tab.~\ref{tab:ATTEN_table}, 3D designs consistently exhibit higher effective frequencies than 2D designs.





\subsubsection{Analysis of Memory Access} 
\begin{table*}[]
    \centering
\caption{The hierachical memory access overhead comparisons between 2D and 3D design of spiking MLP accelerator across different array size and bitwidth }
\begin{tabular}{c c c c c c c}
\toprule
\multirow{2}{*}{Array size H$\times$W, weight/synaptic integration bitwidth}   &   \multicolumn{2}{c}{$16\times 128$, 8b/16b}  & \multicolumn{2}{c}{$64\times 16$, 8b/16b} & \multicolumn{2}{c}{$64\times 16$, 4b/12b} \\  \cline{2-7}
                                        & 2D   & 3D         & 2D   & 3D     & 2D   & 3D    \\ 
\midrule
Activation GLB Latency ($ps$) & 24 &  16 &  68  &   19   &  53   & 58               \\ 
Activation GLB Power ($mW$)  & 1.13 & 0.76 & 1.1 & 0.77 & 1.26 & 0.62 \\ 
Weight GLB Latency ($ps$) & 82 & 26 & 77 & 18 & 80 & 42 \\ 
Weight GLB Power ($mW$) & 0.46 & 0.1 & 0.47 & 0.09 & 0.45 & 0.11 \\ 
\midrule
Activation Buffer Latency ($ps$) & 40 & 16 & 40 & 19 & 47 & 58 \\
Activation Buffer Power ($mW$) & 1.92 & 0.52 & 1.66 & 0.27 & 1.64 & 0.24 \\
Weight Buffer Latency ($ps$) & 28 & 26 & 77 & 18 & 80 & 42 \\
Weight Buffer Power ($mW$) & 1.01 & 0.17 & 1.50 & 0.39 & 1.14 & 0.29 \\
\bottomrule
\end{tabular}    
    \label{tab:Mem_Access_MLP}
\end{table*}

We make a hierachical memory access analysis of different memory blocks. In Tab.~\ref{tab:Mem_Access_MLP}, under different design points with different array size and precision, the memory access latency and energy consumption within 3D design is significantly less than 2D. 

\begin{table}[h]
    \centering
    \caption{2D and 3D Average Wire Length.}
    \small
\begin{tabular}{ccccc}
\toprule
                        & \multirow{2}{*}{Bitwidth} & \multirow{2}{*}{Array Size $H\times W$}  & \multicolumn{2}{c}{Aver. Wire Length(\text{\(\mu\)m})} \\ \cline{4-5}
                        &          &                         & 2D   & 3D      \\ \hline
\multirow{2}{*}{MLP}    & 8b/16b  &  $16\times 128$          & 12.7 & 10.8    \\
                        & 4b/12b   &  $16\times 64$           & 11.8 & 9.59    \\ \hline
\multirow{2}{*}{Attention}        & 16b   & $16\times 16$    & 18.6 & 12.3    \\
                                  & 16b   & $16\times 8$     & 12.9 & 11.5    \\ 
\bottomrule
\end{tabular}
\label{tab:aver_wire_length}
\end{table}


\subsubsection{Wire Length Distribution}
\begin{figure}[ht]
    \centering
    \includegraphics[width=0.5\textwidth, clip, trim={3.2cm 4.5cm 8.4cm 2cm}]{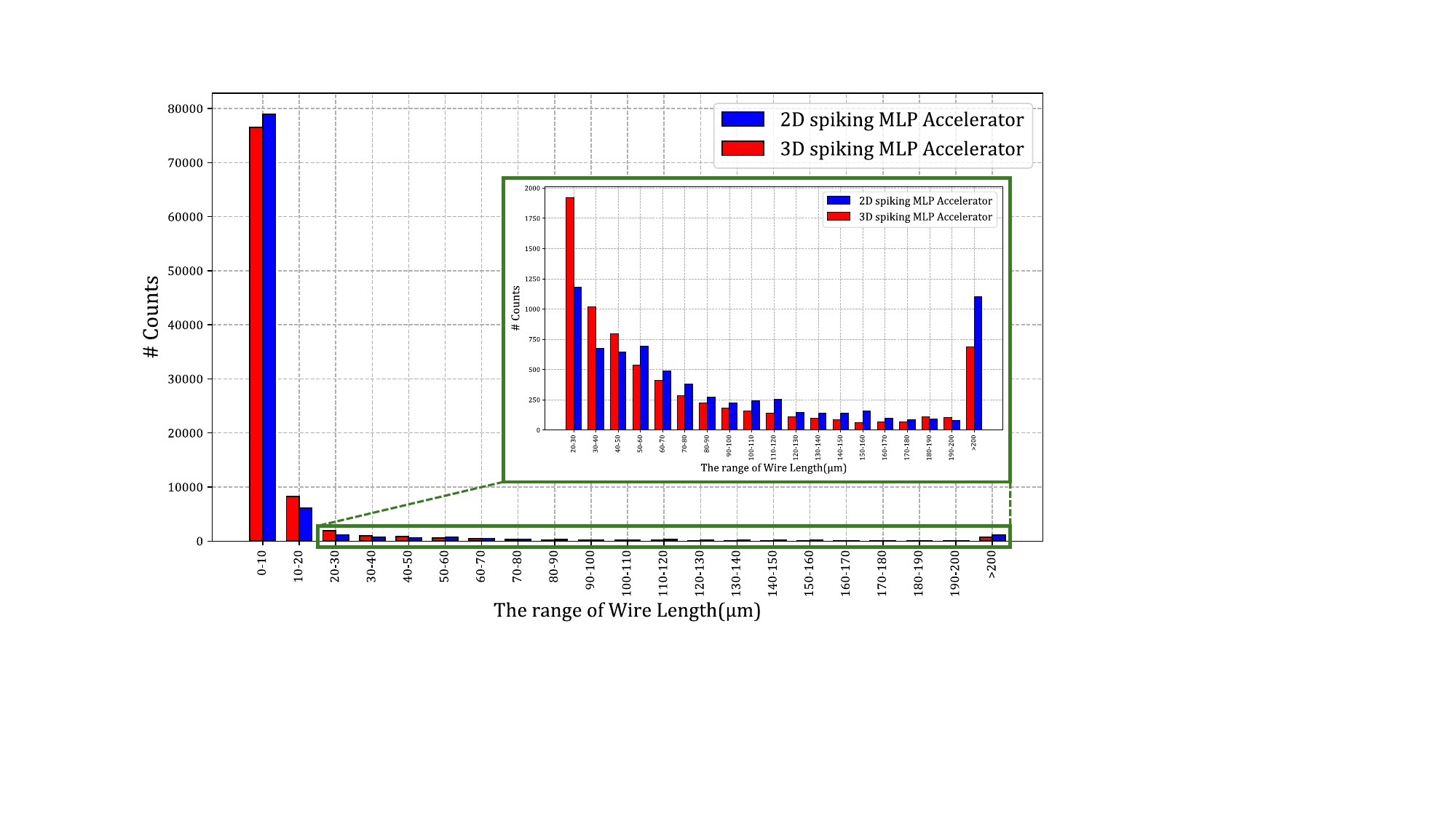}
    \caption{The distribution of Wire Length in the proposed 3D spiking MLP accelerator,}
    \label{fig:wire_length_dist}
\end{figure}

Fig.~\ref{fig:wire_length_dist} illustrates the variance in net wirelength distribution between 2D and 3D IC architectures. In 2D design, the frequency of nets exceeding 50um is consistently higher than in 3D design. This disparity is attributed to the vertical stacking employed in 3D designs, notably between the PE array and spiking generator, which facilitates connections through significantly shorter interconnects. Minimizing wire length is paramount as longer interconnects within a signal path lead to critical paths, adversely affecting the chip's operational speed and overall efficiency.

\section{Conclusion}
In this paper, we introduce the first dedicated 3D accelerator specifically designed for emerging spiking transformers. We identify the spatial and temporal data reuse opportunity on 3D dataflow optimization, and fully exploit this on dedicated 3D accelerators. A tile strategy coupled with kernel fusion is proposed to enable the efficient execution of workloads in spiking transformers. Additionally, our 3D accelerator employs a memory-on-logic and logic-on-logic interconnection scheme via face-to-face (F2F) bonded 3D integration, optimized to minimize energy consumption and latency. Compared to 2D CMOS integration, the 3D accelerator offers substantial improvements. For the spiking MLP workload, it provides a 7.0\% higher effective frequency with 7.8\% less power reduction and 50\% area reduction. The memory access latency and memory access power is reduced by 68.3\% and 69.5\%, respectively. For the spiking self-attention workload, the 3D accelerator is executed at a 6.3\% higher effective frequency, with 50\% area reduction and 1.5\% less power consumption. The memory access latency and memory access power are reduced by 74.2\% and 49.3\%.

\section{Acknowledgement}
This material is based upon work supported by the National Science Foundation under Grants No. 1948201 and No. 2310170 and work supported by the Ministry of Trade, Industry \& Energy of South Korea (1415187652, RS-2023-00234159) and the National Science Foundation under CNS-2235398.

\bibliographystyle{ACM-Reference-Format}
\bibliography{plbib, ref}

\end{document}